\documentclass[sigconf]{acmart}
\usepackage{subcaption}
\usepackage{diagbox} 
\usepackage{stmaryrd}
\usepackage{hyperref}
\usepackage[ruled]{algorithm2e}

\usepackage{xfrac}

\hypersetup{
    colorlinks=true,
    linkcolor=blue,
    filecolor=magenta,      
    urlcolor=cyan,
}
\AtBeginDocument{%
  \providecommand\BibTeX{{%
    \normalfont B\kern-0.5em{\scshape i\kern-0.25em b}\kern-0.8em\TeX}}}

\copyrightyear{2022}
\acmYear{2022}
\setcopyright{acmcopyright}\acmConference[MM '22]{Proceedings of the 30th ACM International Conference on Multimedia}{October 10--14, 2022}{Lisboa, Portugal}
\acmBooktitle{Proceedings of the 30th ACM International Conference on Multimedia (MM '22), October 10--14, 2022, Lisboa, Portugal}
\acmPrice{15.00}
\acmDOI{10.1145/3503161.3547908}
\acmISBN{978-1-4503-9203-7/22/10}

\acmSubmissionID{659}

\begin{document}

\title{Long-term Leap Attention, Short-term Periodic Shift \\for Video Classification}

\author{Hao Zhang}
\affiliation{%
  \institution{Singapore Management University}
  \country{Singapore}
}
\email{hzhang@smu.edu.sg}

\author{Lechao Cheng}
\affiliation{%
  \institution{Zhejiang Lab}
  \city{Hangzhou}
  \country{China}
}
\email{chenglc@zhejianglab.com}

\author{Yanbin Hao}
\authornote{Yanbin Hao is the corresponding author.}
\affiliation{%
  \institution{University of Science and Technology of China}
  \city{Hefei}
  \country{China}}
\email{haoyanbin@hotmail.com}

\author{Chong-wah Ngo}
\affiliation{%
  \institution{Singapore Management University}
  \country{Singapore}
}
\email{cwngo@smu.edu.sg}

\renewcommand{\shortauthors}{Hao Zhang, Lechao Cheng, Yanbin Hao \& Chong-wah Ngo}

\begin{abstract}
 Video transformer naturally incurs a heavier computation burden than a static vision transformer, as the former processes $T$ times longer sequence than the latter under the current attention of quadratic complexity $(T^2N^2)$. The existing works treat the temporal axis as a simple extension of spatial axes, focusing on shortening the spatio-temporal sequence by either generic pooling or local windowing without utilizing temporal redundancy. 
 
 However, videos naturally contain redundant information between neighboring frames; thereby, we could potentially suppress attention on visually similar frames in a dilated manner. Based on this hypothesis, we propose the LAPS, a long-term ``\textbf{\textit{Leap Attention}}'' (LA), short-term ``\textbf{\textit{Periodic Shift}}'' (\textit{P}-Shift) module for video transformers, with $(2TN^2)$ complexity. Specifically, the ``LA'' groups long-term frames into pairs, then refactors each discrete pair via attention. The ``\textit{P}-Shift'' exchanges features between temporal neighbors to confront the loss of short-term dynamics. By replacing a vanilla 2D attention with the LAPS, we could adapt a static transformer into a video one, with zero extra parameters and neglectable computation overhead ($\sim$2.6\%). Experiments on the standard Kinetics-400 benchmark demonstrate that our LAPS transformer could achieve competitive performances in terms of accuracy, FLOPs, and Params among CNN and transformer SOTAs. We open-source our project in \sloppy \href{https://github.com/VideoNetworks/LAPS-transformer}{\textit{\color{magenta}{https://github.com/VideoNetworks/LAPS-transformer}}} .
 \begin{figure}[ht!]
	\centering
	\begin{subfigure}[t]{0.225\textwidth}
		\includegraphics[width=\textwidth]{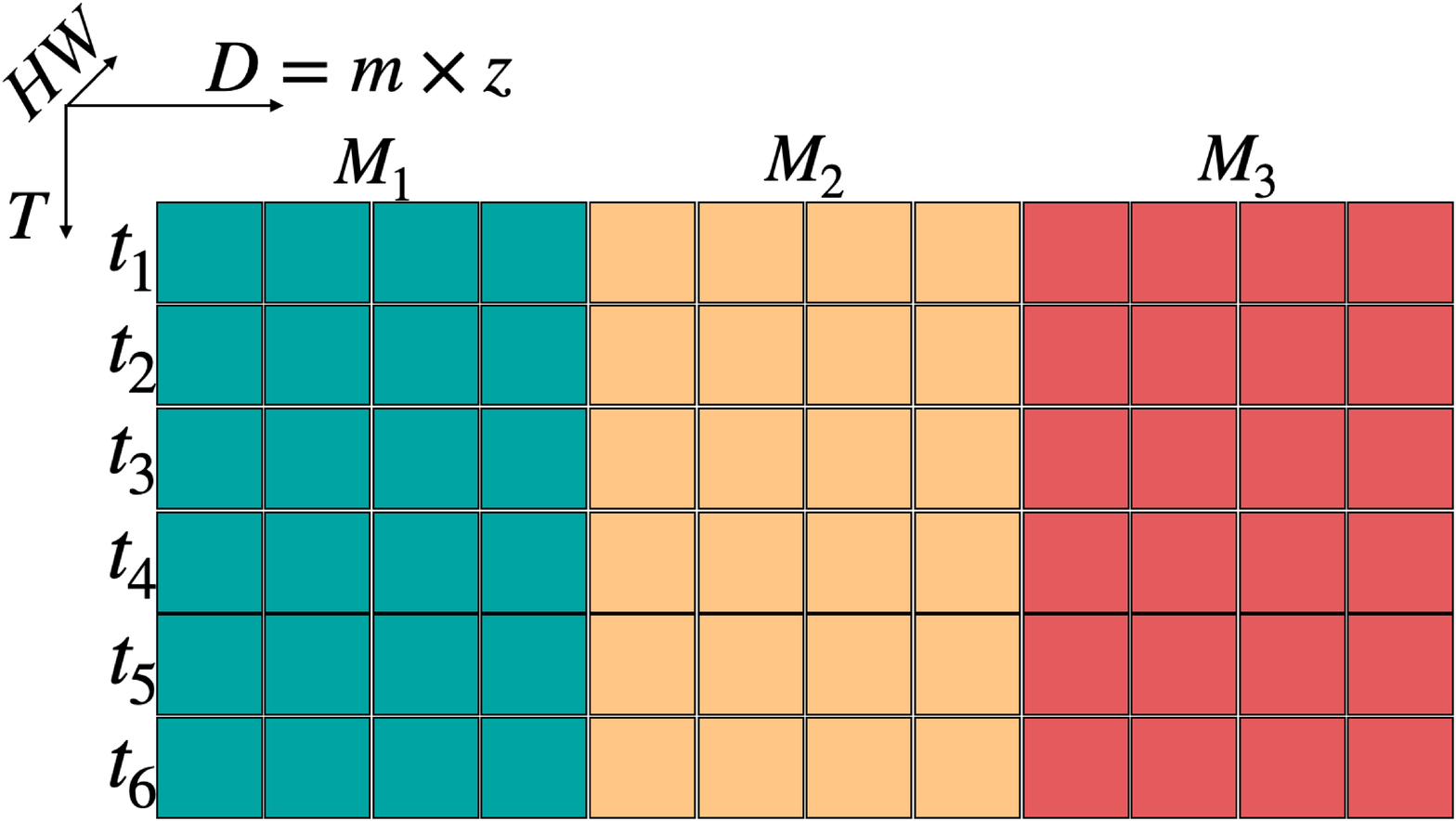}
		\caption{Video Tensor}
		\label{fig:tensor}
	\end{subfigure}
	~ 
	\begin{subfigure}[t]{0.225\textwidth}
		\includegraphics[width=\textwidth]{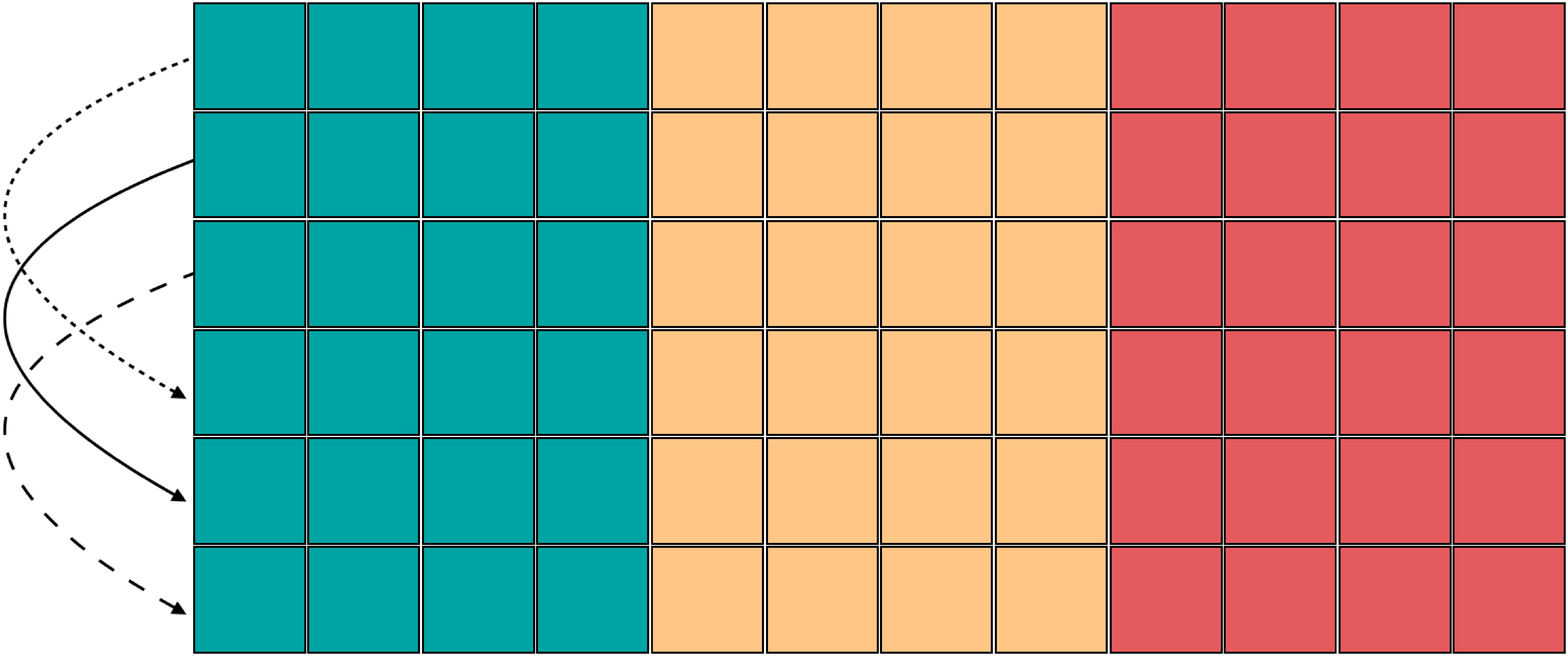}
		\caption{\textbf{Leap Attention}}
		\label{fig:tensor_spa1}
	\end{subfigure}
	\\
	\begin{subfigure}[t]{0.225\textwidth}
		\includegraphics[width=\textwidth]{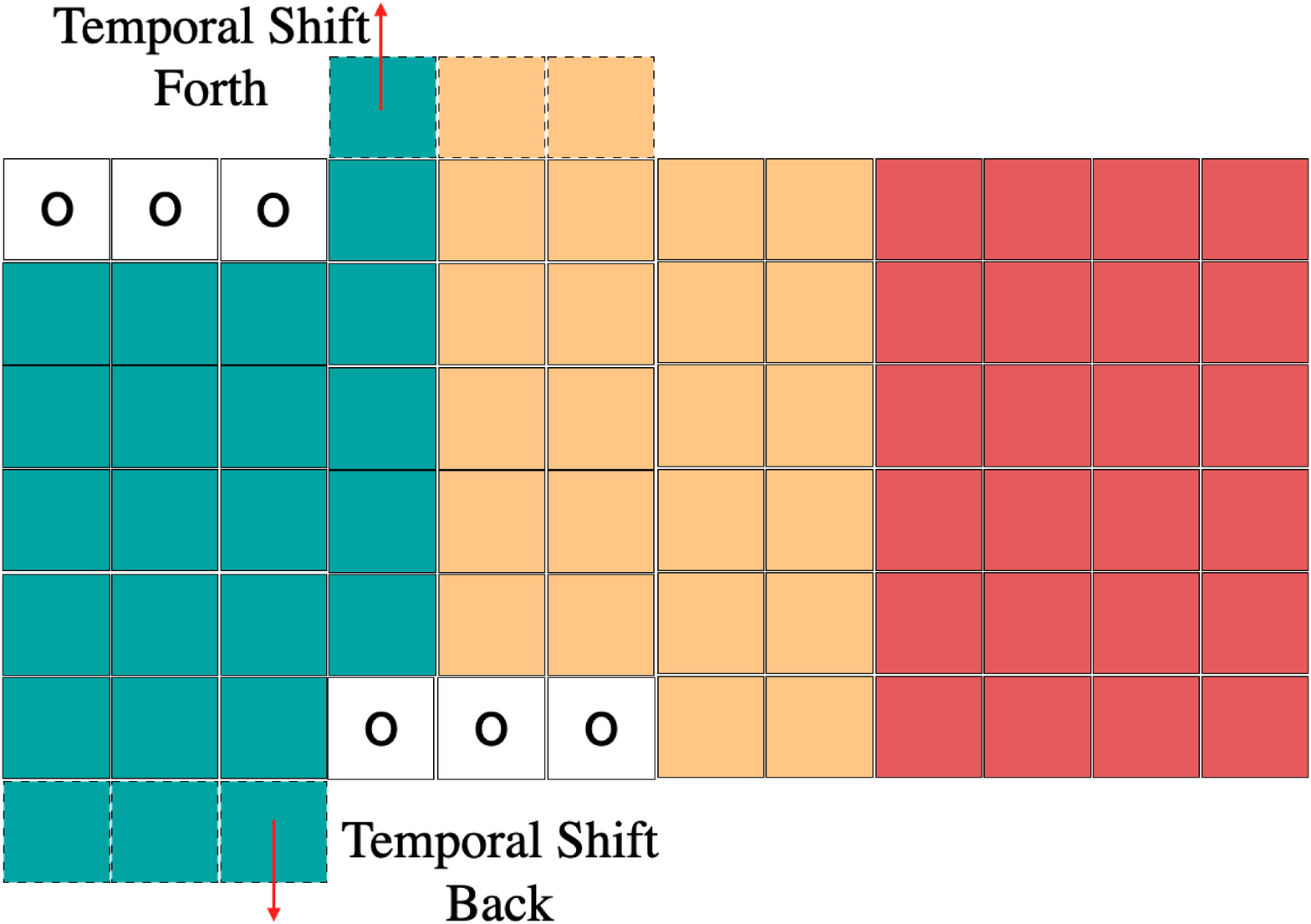}
		\caption{{Plain Temporal Shift}}
		\label{fig:tensor_spa2}
	\end{subfigure}
	\begin{subfigure}[t]{0.225\textwidth}
		\includegraphics[width=\textwidth]{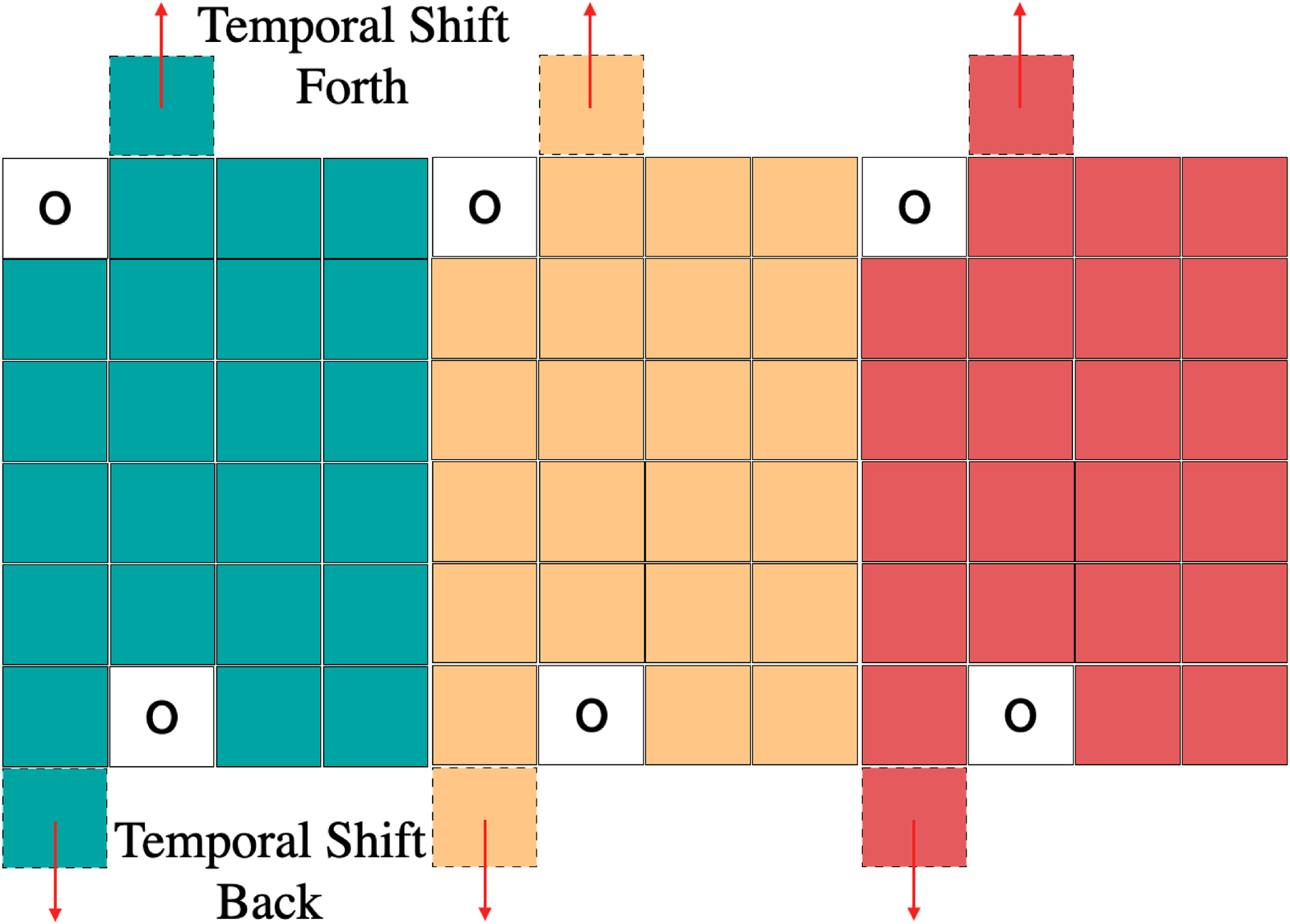}
		\caption{\textbf{Periodic Shift}}
		\label{fig:tensor_spa3}
	\end{subfigure}
	
	\caption{Manipulations on tensor. (a) \textbf{None}; (b) \textbf{Leap Attention}: pairing frames with long-term temporal steps; (c). \textbf{Plain Temporal Shift}; (d) \textbf{Periodic Shift}:  shifting features with periodicity.  $D,m,z,\boxcircle$ denotes total dim, number of heads, dim per head and zeros padding. Features of same hue originates from the same attention head $\boldsymbol{M}_{\{1,2,3\}}$. (The figure is best viewed in color)
	}\label{fig:all_shift}
\end{figure}
\end{abstract}








\maketitle

\section{Introduction}
Transformers recently swept research areas such as NLP \cite{vaswani:attention-2017}, vision \cite{dosovitskiy:image-2020}, and video understanding \cite{bertasius:space-2021} and became the de-facto standards. However, the transformer inherently contains attention mechanism of quadratic complexity, thus innating a disadvantage when dealing with long sequences. This problem is particularly evident with the video input for having $T$\footnote{$T$ refers to the number of frames} longer sequence than the image ($N$), boosting the attention complexity to ($T^2N^2$). 

Current video transformers mainly treat the temporal axis as a simple extension of spatial axes, resorting to generic operators like pooling \cite{fan:multiscale-2021}, local windowing \cite{liu:video-2021}, or factorization of spatio-temporal attention \cite{bertasius:space-2021} to reduce complexity. However, videos are temporally redundant. Temporally neighboring frames are 
generally similar despite being different in micro details. These details gradually accumulate into a qualitative change with enough time evolution. Inspiringly, we suggest avoiding imposing attention on temporally adjacent frames. Alternatively, we could pair frames with a long-term skipped step in between and then apply attention to each pair. This dilated manner would be computationally economical. We also launch a cost-effective variant of the \textit{Shift} operator to capture micro details.

To connect long-term temporal relations, we propose \textbf{\textit{Leap Attention}}. Similar to the dilated convolutions \cite{yu:dilated-2015}, the LA picks frame pairs with discrete temporal distances for attention inputs. As a result, the temporal reception field is expanded at a low computation cost (see Figure \ref{fig:tensor_spa1}). Besides, to equip the LA with multi-scale temporal reception fields, we loop the skipped step in a multiple pyramid way according to layer's depth. We experimentally verify that the LA could surpass the performance of 2D attention and approach 3D attention with an economic complexity ($2TN^2$).

To capture short-term micro variations, we complement the LA with \textbf{\textit{Periodic Shift}}. This \textit{P}-Shift is inspired by the TSM \cite{lin:tsm-2019}, but is customized according to the multi-head mechanism. Recalling that channels of CNN feature are generally sourced from an identical network path, while Multi-Head Self-Attention (MSHA) channels are generated with a latent periodicity. As $m$ heads independently yield sub-channels ($z$-dim) for a total $D$-dim output, every $z$ out of $D$ channels originates from the same head. Our \textit{P}-Shift complies with this characteristics and periodically shifts each  $z$-dim sub-feature of head (Figure \ref{fig:tensor_spa3}). We compare it with a simple copycat named ``plain shift'' (Figure \ref{fig:tensor_spa2}) and observe a better experimental performance.

The complete \textbf{LAPS} module is a cascade of the LA and \textit{P}-Shift as in Figure \ref{fig:atten_all}. It inherits the merit of zero-parameter from both submodules. Thus, it can flexibly convert a static transformer into a video one by replacing the 2D attention. Besides, it could directly load a checkpoint from the 2D pre-training and converge in a few epochs. Extensive experiments on the standard video benchmark verify the efficacy, training/inference efficiency, and flexibility.
Our contributions are summarized below:
\begin{itemize}
    \item \textbf{Leap Attention} adjusts generic spatio-temporal attention into a dilated manner, thus achieving a high cost-effective for video transformers. We also adopt a multi-pyramid strategy to equip the LA with multi-scale temporal reception fields.
    \item \textbf{Periodic Shift} is customized according to the multi-head mechanism, which separately imposes a temporal shift on the output of each attention head. It serves for capturing short-term micro variations between temporal neighbors, with a zero-parameter/FLOPs cost.
    \item \textbf{LAPS (their cascade)} is a zero-parameter, lightweight-FLOPs attention alternative. It can flexibly replace a generic 2D attention and convert a static vision transformer into a video one, facilitating the possibility of developing a video transformer with strong 2D backbones.
\end{itemize}
\section{Related Works}
Our work is relevant to research areas in video and image understandings. Specifically, the prior covers researches in 3D convolutional networks and video transformers; the latter contains recent progress like image transformers. We will review them as below.

\textbf{Convolutional Neural Networks}  for video understanding have been extensively studied and widely applied to video-text pre-training \cite{luo2021coco}, cross-modal analysis \cite{li2021x}, video detection \cite{hao2020person}, ecommerce \cite{cheng2016video, cheng2017video,cheng2017video2shop}, adversarial attack \cite{chen2022attacking, wei2022boosting,wei2022towards,wei2022cross}, interactive search \cite{ma2022reinforcement, wu2021sql}, retrieval \cite{zhang2018fine,han2021fine,li2019w2vv++,wu2020interpretable, han2022adversarial,2022_centerclip,zeng2022point}, hyperlinking \cite{hao2019neighbourhood,hao2021learning,nguyen2017vireo, cheng2017selection}, and caption \cite{tang2021clip4caption,bin2021multi,yuan2020controllable}, in the CNN era; we select and review representative 3D-CNNs as follows. C3D \cite{tran:learning-2015} is a pure 3D-CNN pilot based on a new 3D Conv operator and easily outperforms 2D counterparts on video tasks. Then, the I3D \cite{carreira:quo-2017} increases the depth of network by migrating 3D convs into inception-net \cite{szegedy:going-2015}, and is pre-trained on a large-scale video benchmark named Kinetics to prevent overfitting. To compress parameters and FLOPs, P3D \cite{qiu:learning-2017} factorizes 3D convolution kernel into a combination of 2D spatial + 1D temporal kernels. This factorization demonstrates good efficiency and efficacy, thus gaining popularity in later network design. DB-LSTM \cite{he2021db} learns long-range video sequences with Bi-directional LSTM. To further optimize efficiency, TSM \cite{lin:tsm-2019} adopts the zero-parameter/FLOP Shift operators for temporal modeling. Consequently, it could keep the same complexity\footnote{In terms of model parameters and FLOPs per frame} as the 2D backbone. Furthermore, TIN \cite{shao2020temporal} and Gate-Shift \cite{sudhakaran2020gate} extend temporal shift with adaptive steps, which is predicted by sub-networks. RubiksNet \cite{fanbuch2020rubiks} further reduces computations by proposing a 3D shift operator, which learns 3D steps to shift channels along both spatial and temporal axis. Several works \cite{li2020tea, liu2020teinet, liu2021tam, wang2021tdn, gc2022, tan2021selective} study the impact of long short-term axial contexts on video tasks. They successfully verified that long short-term axial contexts of videos can be utilized in a squeeze-and-excitation manner and are beneficial for content recognition. Small-Big Network \cite{li2020smallbignet} proposes to enlarge the spatial-temporal reception field by paralleling a 3D pooling branch on par with convolutional branches, balancing efficiency and efficacy. Non-Local \cite{wang:non-2018} attempt to implant self-attention into CNNs. Furthermore, CBA-QSA \cite{hao:compact-2020} enhances attention with compact bilinear mapping to emphasize local evidence for sport classification. 
The SlowFast \cite{feichtenhofer:slowfast-2019} and X3D \cite{feichtenhofer:x3d-2020} are SOTAs of 3D-CNN. Specifically, the prior is designed into dual-paths form, one path for semantics and another for motion. The latter relies on searching hyperparameters (e.g., resolution, channels, depth) from a template network, optimizing efficiency and efficacy simultaneously.

\textbf{Image and Video Transformers}. Transformers for images and videos are closely relevant as the latter is often extended from the prior. We elaborate on recent progress for images first, and then turn to videos.

Dosovitskiy et. al \cite{dosovitskiy:image-2020} develops the ViT out of a vanilla language transformer with a modification on accepting sequential patches of an image as input. Since the ViT is computationally heavy, later efforts are devoted to efficiency optimization. For example, Swin \cite{liu:swin-2021} replaces global attention with a local windowed one for reducing the length of tokens. Similar in spirit, the MViT \cite{fan:multiscale-2021} turns to pooling for length shrinking. The rest works develop hybrid models, such as CMT \cite{guo:cmt-2021}, Visformer \cite{chen:visf-2021}, CvT \cite{wu:cvt-2021} and ResT \cite{zhang:rest-2021}, utilizing efficient convolutions for optimization.

Video transformer often contains an extra temporal module than the image one. \cite{bertasius:space-2021} introduces factorization on 3D attention, which decomposes spatio-temporal attention into a combination of spatial and temporal attention as P3D. VTN in \cite{neimark:video-2021} extracts per-frame feature, using 2D-CNN or vision transformer, and then refines these features with a stack of temporal attention encoders. \cite{zhang:token-2021} verify that the Shift operator is still valid for a ViT backbone on video tasks. Moreover, Video Swin \cite{liu:video-2021}, and MViT for video are directly extended from corresponding 2D transformers by inflating 2D windowing/pooling to 3D ones. There is also attention-free pilot, such as MLP-3D \cite{qiu2022mlp} that optimizes efficiency for video understanding.
\section{Method}
\begin{figure*}
	\centering
	\includegraphics[width=0.95\textwidth]{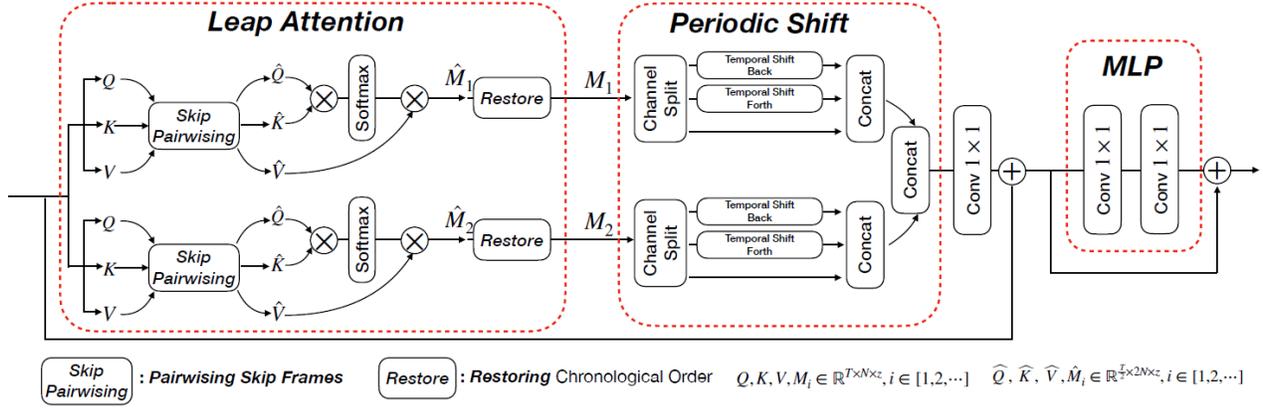}
	~ 
	\caption{An overview of \textbf{LAPS} Encoder. Compared with vanilla Multi-Head Self-Attention, the LAPS contains a \textit{Leap Attention} and \textit{Periodic Shift} module, separately for modeling long-term interaction between frames at a distance and short-term variation within adjacent frames.}
		\label{fig:atten_all}
\end{figure*}

An overview of the \textbf{\textit{Leap Attention with Periodic Shift}} (LAPS) encoder is presented in Figure \ref{fig:atten_all}. The LAPS mainly modifies the vanilla 2D Self-Attention from the following two perspectives: (1). \textit{Leap Attention} for imposing attention on frame pairs with a long-term skipped step; (2). \textit{Periodic Shift} for exchanging features between short-term adjacent frames. We illustrate the detailed pipeline of the LAPS video transformer as below.

A video $\boldsymbol{v} \in \mathbb{R}^{T\times H\times W \times3}$ is gridly divided into a sequential tensor $\boldsymbol{\widetilde{v}}\in \mathbb{R}^{T\times N \times d}$, where $T,H/W,P$ separately denotes clip-length, height/width and patch-size. $N=\frac{H\cdot W}{P^2}$ and $d=P^2\cdot 3$ represents amount of patches (also known as tokens) 
and RGB pixels of a patch.

The sequential tensor $\boldsymbol{\widetilde{v}}$ is then embedded into $D$-dimensional space as in Equation \ref{eq:f1}. Hereby, function $f_{em}(\cdot)$ can be achieved by either a single-layer linear projection or a stack of convolutions. Notably, we follow the trend in \cite{chen:visf-2021} and \cite{wu:cvt-2021} whom adopt a stack of convolutions for  $f_{em}(\cdot)$ and remove extra [{\tt Class}] token on vision tasks. 

\begin{align}
    \boldsymbol{X}_{0}&=f_{em}(\boldsymbol{\widetilde{v}})\label{eq:f1}\\
    \boldsymbol{X}_{0}&\in \mathbb{R}^{T\times N \times D} \nonumber
\end{align}

The video tensor $\boldsymbol{X}_{0}$ is then fed through several encoders for representation learning.

\subsection{LAPS Encoder}
Our video transformer consists of $L$ identical LAPS encoders, each encoder (Figure \ref{fig:atten_all}) contains a LAPS module for spatial-temporal relation modeling and a Multi-Layer Perceptron (MLP) for feature mapping. The function $f_{laps}(\cdot)$ and $f_{mlp}(\cdot)$ separately represents the LAPS and MLP module. 
\begin{align}
&\boldsymbol{\widetilde{X}}_{l-1}=f_{laps}(\boldsymbol{X}_{l-1})+\boldsymbol{X}_{l-1}\label{eq:sacs}\\
&\boldsymbol{{X}}_{l}=f_{mlp}(\boldsymbol{\widetilde{X}}_{l-1})+\boldsymbol{\widetilde{X}}_{l-1}\label{eq:mlp}\\
&\boldsymbol{X}_{l/l-1},\boldsymbol{\widetilde{X}}_{l/l-1} \in \mathbb{R}^{T\times N \times D}\nonumber\\
&D=m\times z,\quad l\in[1,2,\cdots, L]\nonumber
\end{align}
Hereby, $D, m, z$ represent the total dimensions of MSHA, number of heads, and dimensions per head.

The \textbf{LAPS} module $f_{laps}(\cdot)$ contains two sub-modules: Leap Attention (LA) and Periodic Shift (\textit{P}-Shift). The prior serves to model long-term temporal relations between paired frames at a distance; the latter servers to capture short-term variation between adjacent frames. We leave the detailed presentation of LA and \textit{P}-Shift in Section \ref{sec:spa} \& \ref{sec:cs}.


The \textbf{MLP} shares a similar process as it in the vanilla ViT. It contains two layers of linear projection with a GELU activation in between. For simplicity, we implement linear projection in ({\tt Conv-1$\times$1}) form.
\begin{align}
f_{mlp}(\boldsymbol{\widetilde{X}}_{l})=Conv\left(GELU\left(Conv\left(\boldsymbol{\widetilde{X}}_l\right)\right)\right)\label{eq:mlp2}
\end{align}

The \textbf{Prediciton Layer} works on the output $\boldsymbol{\widetilde{X}}_L\in \mathbb{R}^{T\times N\times D}$ from the last encoder. We firstly apply spatial average pooling for per-frame prediction, then average scores across timeline for clip-level prediction, hereby, $FC(\cdot)$ denotes linear projection of shape ``$D\times$ Categories''.
\begin{align}
&\boldsymbol{c}=SpatialAvgPool(\boldsymbol{\widetilde{X}}_L)\\
&\boldsymbol{y}=\frac{1}{T}\sum^{T}_{i=1}{FC(\boldsymbol{c})}\label{eq:vid_pred}\\
&\boldsymbol{y}\in\mathbb{R}^{Categories},\quad \boldsymbol{c}\in\mathbb{R}^{T\times D}\nonumber
\end{align}

\subsection{Leap Attention}\label{sec:spa}
The Leap Attention module is inspired by the success of \textit{Dilated Convolution} \cite{yu:dilated-2015} in the CNN era. Common 2/3D transformers impose attention on continuous receptive field; we propose to compute self-attention on discrete temporal frame pairs. With the dilated strategy, we could enlarge the temporal receptive field of the attention with a neglectable computational overhead.

The \textbf{LA} contains two phases: Grouping frames into pairs and Imposing attention on each pair. We firstly illustrate the strategy of skipped pariwising as below.

Pick the $i$-th sub-tensor $\boldsymbol{M}_i \in \mathbb{R}^{T\times N\times z}$, which would be fed into the $i$-th attention head, out of input tensor $\boldsymbol{X}\in \mathbb{R}^{T\times N \times (m\times z)}$. Hereby, $m$ denotes the total number of attention heads. The $\boldsymbol{Q, K, V}$ tensors are linear projected from $\boldsymbol{M}_i$ as below:
\begin{align}
&\boldsymbol{Q},\boldsymbol{K}, \boldsymbol{V}= \boldsymbol{M}_i \times[ \boldsymbol{W_q}, \boldsymbol{W_k}, \boldsymbol{W_v}]   \\
&\boldsymbol{Q,K,V}\in \mathbb{R}^{T\times N \times z},\quad \boldsymbol{W_{q,k,v}}\in\mathbb{R}^{z\times z}\nonumber
\end{align}

\begin{figure}
	\centering
	\begin{subfigure}[t]{0.10\textheight}
		\includegraphics[width=\textwidth]{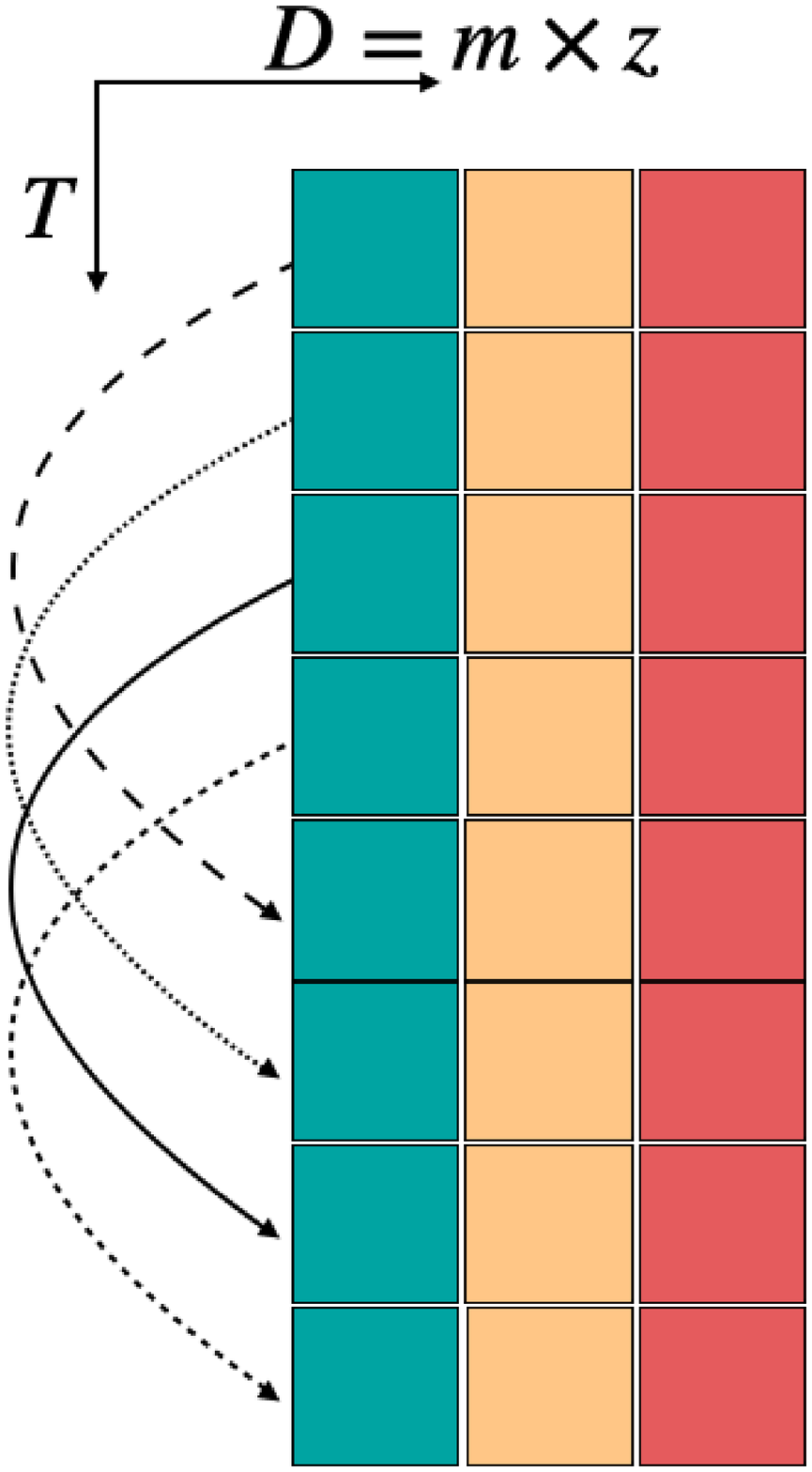}
		\caption{$R=1$}
		\label{fig:p1}
	\end{subfigure}
	~ 
	\begin{subfigure}[t]{0.081\textheight}
		\includegraphics[width=\textwidth]{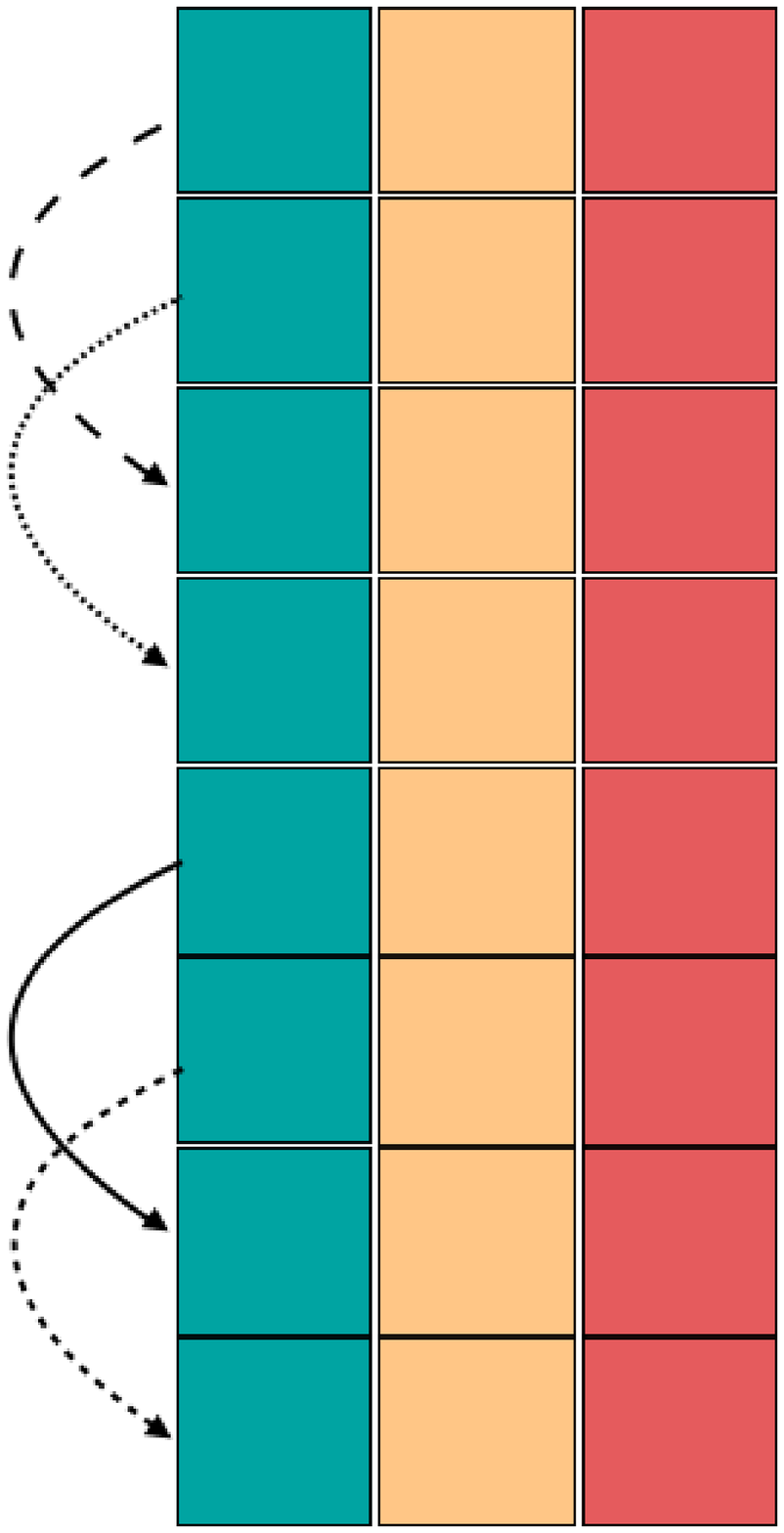}
		\caption{$R=2$}
		\label{fig:p2}
	\end{subfigure}
	\begin{subfigure}[t]{0.077\textheight}
		\includegraphics[width=\textwidth]{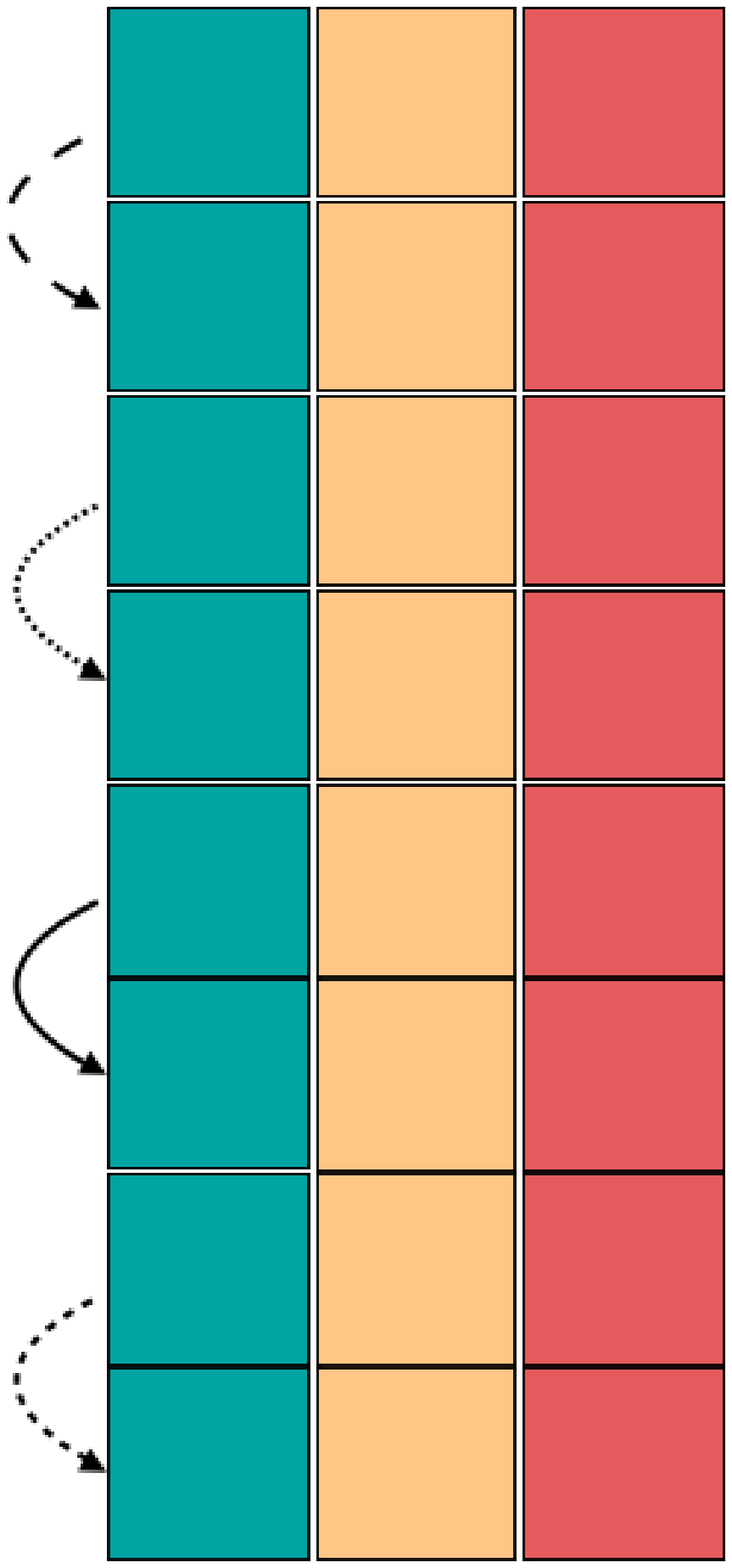}
		\caption{$R=3$}
		\label{fig:p3}
	\end{subfigure}
	\caption{Given a video tensor of $T$ frames, the skip steps of pyramid $R=1/2/3$ range from: [$\frac{T}{2},\frac{T}{4},\frac{T}{8}$]. (The figure is best viewed in color)} 
\vspace{-0.8cm}
\end{figure}

We need two temporal index lists $A$ and $B$ to align frames into pairs. 
Specifically, we separate the temporal indexes $\left[0, 1,2,\cdots,T-1\right]$ into two sorted lists $A$ and $B$, under a condition:
\begin{align}
B-A=\begin{matrix}\left[\underbrace{S, S, \cdots, S}  \right] \\ \frac{T}{2}\end{matrix}
\end{align}
Frames with temporal index ($A[j]$, $B[j]$) are paired, where $j=[0,1,\cdots,\frac{T}{2}]$. Hereby, $S$ denotes the temporal skipped step. For example in Figure \ref{fig:p1}, when $T$=$8$ and $S$=$4$, then, $A$=$[0,1,2,3]$, $B$=$[4,5,6,7]$, the alignment of pairs will be: (0,4), (1,5), (2,6), (3,7). 

We further split $\boldsymbol{Q,K,V}$ into two sub-tensors according to temporal indexes $A$ and $B$:
\begin{align}
&\boldsymbol{Q_{\{A\,\text{or}\, B\}},K_{\{A\,\text{or}\,B\}},V_{\{A\,\text{or}\,B\}}}\in \mathbb{R}^{\frac{T}{2}\times N \times z}\nonumber
\end{align}

The $\boldsymbol{Q_A,K_A,V_A}$ and $\boldsymbol{Q_B,K_B,V_B}$ are concatenated along the 2-nd axis to align frame $t$ and $t+S$ into pairs.
\begin{align}
&\boldsymbol{\widehat{Q}}=Concat\left(\boldsymbol{Q_A},\boldsymbol{Q_B}\right)\\
&\boldsymbol{\widehat{K}}=Concat\left(\boldsymbol{K_A},\boldsymbol{K_B}\right)\\
&\boldsymbol{\widehat{V}}=Concat\left(\boldsymbol{V_A},\boldsymbol{V_B}\right)\\
&\boldsymbol{\widehat{Q}},\boldsymbol{\widehat{K}},\boldsymbol{\widehat{V}}\in \mathbb{R}^{\frac{T}{2}\times ({\color{red}2N}) \times z}\nonumber
\end{align}

We then perform attention on each frame pair (along the 2-nd {\color{red}red} axis), which contains $2N$ tokens:

\begin{align}
&\boldsymbol{\widehat{M}}_i=Softmax\left(\frac{\boldsymbol{\widehat{Q}},\boldsymbol{\widehat{K}}}{\sqrt{z}}\right)\boldsymbol{\widehat{V}}
\end{align}

We reverse the skip pairwsing procedure and restore the $\boldsymbol{\widehat{M}}_i$ to original chronological temporal order $\boldsymbol{\widetilde{M}}_i$ based on index lists $A$ and $B$:
\begin{align}
&\boldsymbol{\widehat{M}}_i\in \mathbb{R}^{\frac{T}{2}\times(2N)\times z}\rightarrow\boldsymbol{\widetilde{M}}_i\in \mathbb{R}^{T\times N\times z}
\end{align}
The outputs from $m$ heads are concatenated as $\boldsymbol{\widetilde{X}}$: 
\begin{align}
&\boldsymbol{\widetilde{X}}=Concat(\boldsymbol{\widetilde{M}}_1,\boldsymbol{\widetilde{M}}_2,\cdots, \boldsymbol{\widetilde{M}}_m)\label{eq:concat}\\
&\boldsymbol{\widetilde{X}}\in\mathbb{R}^{T\times N \times D},\quad D=m\times z\nonumber
\end{align}

The \textbf{Pyramid Skipping} aims to connect frames with various distances into pairs, therefore facilitating the LA to have multi-scale temporal receptive fields. Specifically, we define the skip step $S$ by Equation (\ref{eq:step}) concerning pyramid level $R$. 
\begin{align}
&S=\frac{T}{2^R}
\label{eq:step}
\end{align}
Alignment of temporal pairs with respect to pyramid level $R=1,2,3$ is shown in Figure \ref{fig:p1}-\ref{fig:p3}. Since a transformer contains $L$ amounts of LAs (residing in each encoder), we loop pyramid level $R$ by values $[1,2,3]$. We also present the complete pipeline of ``leap attention'' in Algorithm \ref{alg:algorithm}. 

\begin{algorithm}[t]
    \caption{Leap Attention}
    \label{alg:algorithm}
    \LinesNumbered
    \KwIn {video tensor $\boldsymbol{X}\in \mathbb{R}^{T\times N \times (m\times z)}$, pyramid-level $R$} 
    \KwOut {video tensor $\boldsymbol{\widetilde{X}}\in \mathbb{R}^{T\times N \times (m\times z)}$ refactor from skipped neighbors}
   Let temporal skip step $S=\frac{T}{2^{R}}$\\
    \For{i = 1 to m}{  
    Select the $i$-th head $\boldsymbol{M}_i\in \mathbb{R}^{T\times N\times z}$ from $\boldsymbol{X}$.\\
    Generate $\boldsymbol{Q},\boldsymbol{K},\boldsymbol{V} \in \mathbb{R}^{T\times N\times z}$ from $\boldsymbol{M}_i$ \\
    Init index list $A$ and $B =[\varnothing$]\\
    \For{t=1 to T}{
    \If{$t\notin \{A\cup B\}$}{Add $t$ to set $A$: $A=A\cup\{t\}$\\
    Add $t+S$ to set $B$: $B=B\cup\{t+S\}$}
    }
    Sort A and B in ascending order\\
    Select sub-tensor by $A$: $\boldsymbol{Q/K/V_A}\in \mathbb{R}^{\frac{T}{2}\times N \times z}$.\\
    Select sub-tensor by $B$: $\boldsymbol{Q/K/V_B}\in \mathbb{R}^{\frac{T}{2}\times N \times z}$.\\
    Concate over $2$-nd axis: $\boldsymbol{\widehat{Q}/\widehat{K}/\widehat{V}}\in \mathbb{R}^{\frac{T}{2}\times (2N) \times z}$\\
    Perform attention over $2$-nd axis ($2N$ tokens):
$\boldsymbol{\widehat{M}}_i=Softmax\left(\frac{\boldsymbol{\widehat{Q}}\boldsymbol{\widehat{K}}}{\sqrt{z}}\right)\boldsymbol{\widehat{V}}$\\
Restore chronological order based on list $A$, $B$: $\boldsymbol{\widehat{M}}_i\in \mathbb{R}^{\frac{T}{2}\times(2N)\times z}\rightarrow\boldsymbol{\widetilde{M}}_i\in \mathbb{R}^{T\times N\times z}$\\
}
    Concate $m$ heads outputs $\boldsymbol{\widetilde{M}}_i$ into  $\boldsymbol{\widetilde{X}}$.\\
    \textbf{return} $\boldsymbol{\widetilde{X}}\in \mathbb{R}^{T\times N \times (m\times z)}$
\end{algorithm}

\textbf{Complexity Analysis.} We theoretically compare complexities between the LA, 2D, and 3D attentions. Supposing that a clip contains $T$ frames, with $N$ tokens per frame, we then separate frames into ${g}$ groups and apply attention to each group. The total complexity will be ${g}\times\left(\frac{T}{g}\times N\right)^2$=$\frac{T^2N^2}{g}$. This complexity will be: (1). $TN^2$ with 2D attention, when $g$=$T$; (2). $2TN^2$ with the LA, when $g$=$\frac{T}{2}$; (3). $T^2N^2$ with 3D attention, when $g$=$1$. Overall, the LA introduces a small overhead to 2D attention while being obviously more efficient than 3D attention.



\subsection{Periodic Shift}\label{sec:cs}
The \textit{P}-Shift is inspired by the Temporal Shift Module (TSM) \cite{lin:tsm-2019}, originally designed for capturing short-term temporal dynamics in CNNs, but specially adjusted according to the multi-head mechanism of the transformer.


Recalling the output of multi-head attention is $\boldsymbol{\widetilde{X}}\in \mathbb{R}^{T\times N \times D}$ (Equation (\ref{eq:concat})), where $D$-dimentional feature is concatenated from $m$ amount of $z$-dimentional sub-feature, namely $D=m\times z$.


\textbf{Periodic Shift (\textit{P}-Shift)} uniformly processes the sub-feature from each head. Specifically, $D$ channels are firstly decomposed into $m\times z$ shape; then each $z$-channels are split into three $\boldsymbol{p}^{a'},\boldsymbol{p}^{b'},\boldsymbol{p}^{c'}$ parts:

\begin{align}
&\boldsymbol{\widetilde{X}} = \left[\boldsymbol{{p}}^{a'},  \boldsymbol{{p}}^{b'}, \boldsymbol{{p}}^{c'}\right]\nonumber\\
&\boldsymbol{p}^{\{a',b',c'\}}\in \mathbb{R}^{T\times N \times m\times \{a',b',c'\}}\nonumber\\
&a' + b' + c' = z\nonumber
\end{align}

The shift operation is conducted on of $\boldsymbol{p}^{a'},\boldsymbol{p}^{b'}$ part, while keeping $\boldsymbol{p}^{c'}$ unchanged (Figure \ref{fig:tensor_spa3}). The $t,i,u,j$ denotes indices for temporal, spatial, attention head, and dim (per head) axis. The ``$\Leftarrow$'' denotes assignment operation.

\begin{align}
&\boldsymbol{p}^{a'}_{\left(t, i, u, j\right)}\Leftarrow\boldsymbol{p}^{a'}_{(t-1, i, u, j)},\quad0\leq j\leq a' \\
&\boldsymbol{p}^{b'}_{\left(t, i, u, j\right)}\Leftarrow\boldsymbol{p}^{b'}_{(t+1, i, u, j)},\quad0\leq j\leq b' \\
&\boldsymbol{p}^{c'}_{\left(t, i, u, j\right)}\Leftarrow\boldsymbol{p}^{c'}_{(t, i, u, j)},\quad0\leq j\leq c'\\
&0\leq t\leq T,\quad0\leq i\leq N,\quad0\leq u\leq m\nonumber
\end{align}

\textbf{Plain Temporal Shift} directly inherits the original TSM by splitting $D$ channels into three parts:
\begin{align}
&\boldsymbol{\widetilde{X}} = \left[\boldsymbol{{p}}^{a},  \boldsymbol{{p}}^{b}, \boldsymbol{{p}}^{c}\right]\nonumber\\
&\boldsymbol{p}^{\{a,b,c\}}\in \mathbb{R}^{T\times N \times \{a,b,c\}}\nonumber\\
&a + b + c = D\nonumber
\end{align}
where the split $\boldsymbol{p}^{a}$/$\boldsymbol{p}^{b}$ are shift with prior/post time-stamps along the temporal axis. The split $\boldsymbol{p}^{c}$ remains unchanged as in Figure \ref{fig:tensor_spa2}. 

Figure \ref{fig:all_shift} shows that the {\textit{P}-Shift} is less biased than the plain shift in manipulating features from different heads, thus resulting in better performances.
\section{Experiments}
We evaluate the LAPS-Former on the below standard benchmarks in terms of Top-1/5 accuracy, GFLOPs, Params.

\textbf{Kinetics-400} \cite{carreira:quo-2017} is a large-scale benchmark for video classification. It defines 400 categories on $\sim$246k/20k training/validating clips, with 10 seconds per clip.

\textbf{UCF-101} \cite{soomro:ucf101-2012}  serves as a small-scale datasets for classifying ``human-object'' interactions and sports. It contains three splits on 101 categories. Follow prior works, we report performance on the split 1, with 9,537/3,783 train/validation clips. The average clip-length is 5.8 seconds.

\textbf{EGTEA Gaze+} \cite{li2018eye} is a First-Person-View video dataset captured by wearable-device in daily life. It covers 106 action categories defined on 3.1 seconds long video clips. We adopt the split 1, which contains 8,299/2,022 training/validation clips, for evaluation and comparison.

\subsection{Settings}
We implant the LAPS module into Visformer \cite{chen:visf-2021} which is pre-trained on ImageNet data with 10k categories. All experiments follow the same recipe unless particularly specified. For experiments with variable resolutions and clip-lengths, we linearly scale the lr according to batch-size as in \cite{goyal:accurate-2017}.


\textbf{Training}. Each clip contains 8 frames of $224\times224$ (320 for high resolutions) shape, with a sampling rate of 8. Augmentations, such as random resizing, horizontal flipping, and color perturbation (e.g., brightness, saturation, and hue), are adopted to reduce overfitting. The training schedule is set to 18 epochs, decaying lr decayed by 0.1 at the 10/15-th epoch. The shift ratio is fixed to $\frac{a}{D}$, $\frac{b}{D},\frac{a'}{z}$, $\frac{b'}{z}$=$\frac{1}{8}$ as in TSM.   

\textbf{Inference}. We uniformly sample 5 clips from a video, resizing their short-sides to 224. We then crop three squared sub-clips from each clip's ``left, center, right'' parts. The prediction of a video is averaged over those of 15 sub-clips.

\subsection{Ablation Study}
We experimentally study impacts of various temporal operations (e.g, Plain Shift, $P$-Shift and Leap Attention), pyramid combinations, backbones, clip lengths, and resolutions on the LAPS module.

\textbf{Leap Attention} \textit{vs} \textbf{2/3D Attention}. We implement different attentions, such as generic 2/3D and Leap Attentions, in the MSHA. As in Table \ref{tab:tab1}, the LA easily surpasses 2D attention (75.84\% \textit{vs} 74.00\%), with only 2.6\% computational overhead. Though the 3D attention (76.31\%) performs slightly better than LA, it suffers from a increament of 18.9\% computations. Moreover, combining the LA with \textit{P}-Shift (76.04\%) approaches the performance of 3D attention, with much less costs. 

\textbf{Periodic Shift} \textit{vs} \textbf{plain Shift}. To eliminate the influence of attention, both models adopt simple 2D attention. As is shown in Table \ref{tab:tab1}, both Shifts (74.86\%, 75.19\%) outperform 2D Baseline, verifying the efficacy of implanting the Shift into the attention. Besides, the \textit{P}-Shift performs better than the plain counterpart (75.19\% \textit{vs} 74.86\%), showing that output from each head requires to be equally shiftted.

\begin{table}
\centering
\small
\begin{tabular}{lllrr}
	\toprule
	Model  &MSHA&GFLOPs& Params&Top-1 \\
	&&&(M)&(\%)\\
	\midrule
	Base2D &2D Atten &39.1&39.8&74.00\\
	Base3D &3D Atten &{46.5}  ($18.9\%\uparrow$)&39.8&{76.31}\\
	\midrule
	 &  Plain Shift & 39.1&39.8&74.86\\
	 LAPS&  \textit{P}-Shift & 39.1&39.8&75.19\\
	 &  LA &40.1 ($2.6\%\uparrow$)&39.8&75.84\\
	& \textit{P}-Shift + LA & 40.1 ($2.6\%\uparrow$)&39.8&76.04\\

	\bottomrule
\end{tabular}
\caption{Impacts of temporal modules on \textit{LAPS-Former} and Comparisons with 2/3D Attention baselines.}
\vspace{-0.3cm}
\label{tab:tab1}
\end{table}

Notably,  our LAPS-Former shares the same amount of parameters as the 2D Baseline, verifying that the LAPS module alone is a zero-parameter operator.



\textbf{Pyramid combinations}. We test LAPS with various combinations of skipped steps. Since our LAPS transformer contains eight LAs, we can 
set an identical or different $S$ for them. For multi-pyramid settings, we loop $R$ by values $[1,2,3]$ with respect to the depth of LA. Additionally, a smaller $R$ value indicates a larger skipped step $S$ between paired frames (Equation (\ref{eq:step})).


\begin{table}
\centering
\small
\begin{tabular}{llcr}
	\toprule
	Model  &Pyramids&Skipped Steps&Top-1 \\
	\midrule
	LAPS&R=$[3,3,3]$&S=$[\sfrac{1}{8},\sfrac{1}{8},\sfrac{1}{8}]\cdot T$&75.55\\
& R=$[2,2,2]$&S=$[\sfrac{1}{4},\sfrac{1}{4},\sfrac{1}{4}]\cdot T$&75.82\\
& R=$[1,1,1]$&S=$[\sfrac{1}{2},\sfrac{1}{2},\sfrac{1}{2}]\cdot T$&75.86\\
	&R=$[1,2,3]$&S=$[\sfrac{1}{2},\sfrac{1}{4},\sfrac{1}{8}]\cdot T$&\textbf{76.04}\\
	\bottomrule
\end{tabular}
\caption{Comparisons of \textit{LAPS-Former} under different combinatons of skipped steps (pyramids). (Top-1 Accuracy).}
\vspace{-0.8cm}
\label{tab:tab2}
\end{table}

Table \ref{tab:tab2} compares different combinations. We observe that: (1). The LAs with multiple pyramids (76.04\%) outperform single-level ones. A potential reason is that multi-scale temporal information is complementary; (2). Among all single levels, precision is positively correlated with $S$, (when $\frac{T}{8}\rightarrow \frac{T}{4} \rightarrow \frac{T}{2}$, then $75.55\%\rightarrow 75.82\% \rightarrow 75.86\%$), showing that LA prefers intaking distinct features than similar ones.

\textbf{Length of tokens}. We study the impact of tokens' length on LAPS-Former. Factors like clip length and spatial resolution determine this length. 

\begin{table}[h!t]
\begin{center}
	\small
	\begin{tabular}{llrrrr}
		\toprule
		Model &Base&\#F$\times$Res &GFLOPs &Paras & Top-1\\
		&&\textit{T}$\times$\textit{HW}  & &(M) &(\%) \\
		\midrule
		Base2D & Visf&8$\times$$224^2$ &39.1&39.8&74.00\\
		LAPS & Visf&8$\times$$224^2$ &40.1&39.8&76.04\\
		LAPS (B)& Visf&8$\times$$320^2$ &86.4&40.0&77.56\\
		LAPS (L) & Visf&16$\times$$320^2$ &173.0&40.0&78.71\\ 
		LAPS (H) & Visf&32$\times$$320^2$ &346.0&40.0&79.72\\ 
		LAPS (E) & Visf&32$\times$$360^2$ &434.0&40.2&80.03\\
		\midrule
		Base2D & RstS&8$\times$$224^2$ &16.9&13.4&68.13\\
		LAPS (RS) & RstS&8$\times$$224^2$ &17.6&13.4&69.72\\
		LAPS (RB) & RstB&8$\times$$224^2$ &39.3&29.9&71.89\\
		LAPS (RL)& RstL&32$\times$$320^2$ &595.0&51.2&76.71\\ 
		\midrule
		Base2D & ViT&8$\times$$224^2$ &141.0&86.1&76.41\\
		LAPS (VT) & ViT&8$\times$$224^2$ &146.0&86.1&78.15\\
		\bottomrule
	\end{tabular}
\end{center}
\caption{Impacts of backbone, clip-length and spatial resolution on \textit{LAPS-Former} (Top-1 Accuracy).}
\vspace{-0.8cm}
\label{tab:tab5}
\end{table}

Table \ref{tab:tab5} shows that LAPS benefits from both large temporal/spatial resolution. Specifically, we observe that: (1) LAPS (B) intakes a larger spatial resolution (224$\rightarrow$320) over origin LAPS, resulting in a boost of 1.52\% top-1 accuracy (76.04\%$\rightarrow$77.56\%); (2) From LAPS (B)-(H),    precision increases (77.56\%$\rightarrow$78.71\%$\rightarrow$79.72\%) with the temporal resolution (8f$\rightarrow$16f$\rightarrow$32f).

\begin{table*}[h!t]
	\begin{center}
		\small
		\begin{tabular}{lllrrrrrr}
			\toprule
			Model &Base& Pretrain &\#F$\times$Res  &GFLOPs$\times$Views & Params& Training &Top-1 & Top-5 \\
			& & &  ($T\times HW$)&&(M)&Epochs&(\%)&(\%) \\
			\midrule
			I3D \cite{carreira:quo-2017} from \cite{feichtenhofer:x3d-2020}&InceptionV1&IN-1K& $250\times224^2$ & $108\times$ NA & 12.0&-&71.10 &90.30\\
			Two-Stream I3D \cite{carreira:quo-2017} from \cite{feichtenhofer:x3d-2020}&InceptionV1&IN-1K& $500\times224^2$ & $216\times$ NA & 25.0 &-&75.70 &92.00\\
			TSM \cite{lin:tsm-2019} &ResNet50&IN-1K& $16\times256^2$ & $65.0\times10$&24.3&100 &74.70&-\\
			S3D-G \cite{xie2018rethinking}&InceptionV1&IN-1K&$250\times224^2$&$71.3\times$NA&11.5&112&74.70&93.40\\
			TEA \cite{li2020tea} &ResNet50&IN-1K& $16\times256^2$ & $70.0\times 30$ & 35.3&50&76.10 &92.50\\
			TEINet \cite{liu2020teinet} &ResNet50&IN-1K& $16\times256^2$ & $66.0\times 30$ & 30.8&100&76.20 &92.50\\
			TANNet \cite{liu2021tam} &ResNet50&IN-1K& $16\times256^2$ & $86.0\times 12$ & 25.6&100&76.90 &92.90\\

			Non-Local R50 \cite{wang:non-2018} from \cite{feichtenhofer:x3d-2020} &ResNet50&IN-1K& $128\times256^2$ & $282.0\times 30$ & 35.3&118&76.50 &92.60\\
			Non-Local R101 \cite{wang:non-2018} from \cite{feichtenhofer:x3d-2020}&ResNet101&IN-1K& $128\times256^2$& $359.0\times 30$ &54.3&196&77.70 &93.30\\
		    Small-Big \cite{li2020smallbignet} &ResNet101&IN-1K& $32\times224^2$& $418.0\times 12$ &-&110&77.40 &93.30\\
			TDN-R50 \cite{wang2021tdn} &ResNet50&IN-1K& $24\times256^2$& $108.0\times 30$ &26.6&100&78.40 &93.60\\

			TDN-R101 \cite{wang2021tdn} &ResNet101&IN-1K& $24\times256^2$& $198.0\times 30$ &43.9&100&79.40 &94.40\\
			GC-TDN-R50 \cite{gc2022} &ResNet50&IN-1K& $24\times256^2$& $110.1\times 30$ &27.4&100&79.60 &94.10\\
			SlowFast $8\times8$ \cite{feichtenhofer:slowfast-2019}& ResNet50&None&$32\times256^2$ &  $65.7\times30$&-&196&77.00&92.60\\
			SlowFast $16\times8$ \cite{feichtenhofer:slowfast-2019}& ResNet101+NL&None&$32\times256^2$  & $234.0\times30$&59.9&196&79.80&93.90\\
			X3D-L \cite{feichtenhofer:x3d-2020}&X2D &None&$16\times356^2$  &$24.8\times30$&6.1&256&77.50&92.90\\
			X3D-XL \cite{feichtenhofer:x3d-2020}&X2D &None&$16\times356^2$  &$48.4\times30$&11.0&256&79.10&93.90\\
			
			
			\midrule
			ViT (Video) \cite{zhang:token-2021}&ViT-B&IN-22K& $8\times224^2$ &$134.7\times30$ &85.9 &18&76.00&92.50 \\
			TokShift \cite{zhang:token-2021}&ViT-B&IN-22K& $16\times224^2$ &$269.5\times30$ &85.9&18 &78.20&93.80 \\
			TokShift (MR) \cite{zhang:token-2021}&ViT-B&IN-22K& $8\times256^2$ &$175.8\times30$&85.9&18& 77.68&93.55\\
			VTN \cite{neimark:video-2021}&ViT-B&IN-22K& $250\times224^2$ &$4218.0\times1$&114.0&25&{78.60}&93.70\\
			TimeSformer \cite{bertasius:space-2021}&ViT-B&IN-22K& $8\times224^2$ &$590.0\times3$&121.4&15&78.00&93.70\\
			Video Swin \cite{liu:video-2021}&Swin-B&IN-1K& $32\times224^2$ &$281.6\times12$&88.0&30&80.60&94.60\\
			MViT \cite{fan:multiscale-2021}&MViT-B&None& $64\times224^2$ &$455.0\times9$&36.64&200&81.20&95.10\\
			\midrule
			LAPS&Visformer&IN-10K& $8\times224^2$ &$40.1\times15$ &39.8& 18&76.04&92.56 \\
			LAPS (L)&Visformer&IN-10K& $16\times320^2$ &$173.0\times15$ &40.0&18 &78.71&93.77 \\
			LAPS (H)&Visformer&IN-10K& $32\times320^2$ &$346.0\times15$ &40.0 &18&79.72&94.08 \\
			LAPS (E)&Visformer&IN-15K& $32\times360^2$ &$434.0\times15$ &40.2 &18&80.03&94.48 \\
			
			\bottomrule
		\end{tabular}
	\end{center}
	\caption{\textbf{Comparison to state-of-the-arts on Kinetics-400 Val}.}
	\vspace{-0.3cm}
	\label{tab:all}
\end{table*}

\textbf{On various backbones}. Our LAPS could be flexibly and directly plugged into backbones with a \textit{standard} multi-head self-attention mechanism. We verify this by implanting it into 2D transformers, such as Visformer \cite{chen:visf-2021}, ResT \cite{zhang:rest-2021} and ViT\cite{dosovitskiy:image-2020}. The ResT and ViT backbone is separately pre-trained on ImageNet-1k and 22k. 

Table \ref{tab:tab5} also lists performance of LAPS on 2D backbones. Our LAPS consistently boosts Baseline-2D ($68.13\%\rightarrow 69.72\%$ for ResT, $76.41\%\rightarrow 78.15\%$ for ViT) while introducing small/zero extra GFLOPs or parameters. Besides, the performance is further improved with more encoders, longer clip-length, and larger resolution.

\subsection{Comparison with the State-of-the-Art}
In Table \ref{tab:all}, we compare LAPS-Former with the current 3D-CNN and video transformer SOTAs in terms of training/inference efficiency (\#Epochs, GFLOPs), model size (Params), precision (Top-1/5 accuracy). 

\textbf{LAPS \textit{vs} 3D-CNNs}. Firstly, like pilot transformers, the LAPS-Former shows a higher training efficiency (10 times fewer epochs) than all 3D-CNNs. An extra factor is that LAPS is built from temporal operators of zero-parameter. Thus, we could directly load model and resume training from pre-trained weights of 2D tasks. Secondly, the LAPS achieves better performance than 3D-CNNs, such as Non-Local (79.72\% \textit{vs} 77.7\%, 346G \textit{vs} 359G) and  TSM (76.04\% \textit{vs} 74.7\%, 40.1G \textit{vs} 65G), with less computations. Finally, our LAPS-Former achieves better performance (80.03\%) compared with strong SOTAs in the CNN era, such as SlowFast (79.8\%), X3D (79.1\%), with slightly more computations. The reason is that optimizations for 3D convolutions have been intensively studied and applied in designing. For instance, the SlowFast adopts 3D convolution decomposition, and the X3D is automatically searched from a base net.

\textbf{LAPS \textit{vs} SOTAs transformer}. Our LAPS-Former (78.71\%, 173G, 40M) surpasses pilot video transformers, such as TokShift (78.2\%, 269.5G, 85.9M),  VTN (78.6\%, 4.2$\times 10^3$G, 114M) and TimeSformer (78.0\%, 590G, 121.4M) under all metrics. Besides, our model (80.03\%) is slightly lower than strong transformers like Video Swin (80.6\%) and MViT (81.2\%). A possible reason might be they directly learn of cube' embeddings in generating video sequences, whereas our LAPS-Former relies on patch embeddings. Besides, our 2D backbone (i.e., Visformer) is weaker than Swin and static MViT. However, the LAPS module still has merits, such as fewer training epochs and being more flexible than Video Swin and MViT. As our LAPS-Former can adapt to various backbones, it could potentially take advantage of newly emerging 2D transformers for 3D modeling.

\subsection{Fine-tune on small-scale datasets}
We further verify the effectiveness of LAPS on downstream small-scale video datasets, such as UCF-101 and EGTEA Gaze+ datasets. We initialize the training phase with pre-trained weight on the Kinetics-400 datasets to save training time and avoid overfitting. Their training/inference recipe is almost identical to the Kinetics-400 datasets, except that the training schedule is separately optimized for both datasets. Specifically, we train 18 epochs, decaying LR by 0.1 at 10/15-th epoch for UCF-101; whereas for EGTEA-Gaze+, the total epoch is set to 36, decaying LR by 0.1 at 20/30-th epoch. Experimental results for UCF-101 and EGTEA-Gaze+ is separately present in Table \ref{tab:tab9} and Table \ref{tab:tab8}.

\begin{figure*}[h!t]
	\centering
	\includegraphics[width=0.80\textwidth]{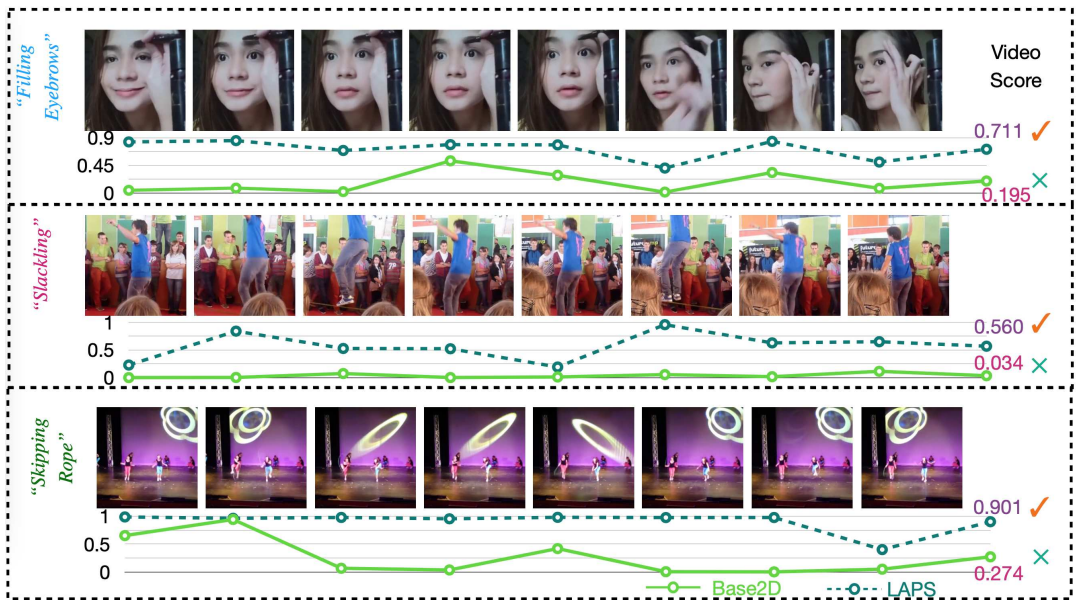}
	~ 
	\caption{Visualization of Video Exemplars. We test Base2D and LAPS models on the validation set of the Kinetics-400 dataset and pick video examples with significant improvement. To highlight the impact of temporal relations, we present per-frame prediction scores and their trend along with clip frames. The solid and dashed line separately denotes predictions from Base2D and LAPS models. (The figure is best viewed in color) }
		\label{fig:vis_all}
\end{figure*}

\begin{table}
\begin{center}
\small
\begin{tabular}{l|lccc}
\toprule
Model &Pretrain& Res & \# Frames &UCF101\\
 & & ($H\times W$) &$T$  & Acc1 (\%)\\

\midrule
I3D \cite{carreira:quo-2017}&K-400 & $224\times 224$ &250 &84.50\\
P3D \cite{qiu:learning-2017}&K-400 & $224\times 224$ &16 &84.20\\
Two-Stream I3D \cite{carreira:quo-2017}&K-400 & $224\times 224$ &500 &93.40\\
TSM \cite{lin2019tsm}&K-400 & $256\times 256$ &8 &95.90\\
\midrule
ViT (Video) \cite{zhang:token-2021}&IN-22K& $256\times256$ & 8 &91.46\\
TokShift \cite{zhang:token-2021}&K-400& $256\times256$ & 8&95.35\\
TokShift-L (HR) \cite{zhang:token-2021}&K-400& $384\times384$ & 8&{96.80}\\
LAPS&K-400& $224\times224$ & 8&95.43\\
LAPS (H)&K-400& $320\times320$ & 32&\textbf{96.85}\\
\bottomrule
\end{tabular}
\end{center}
\caption{Fine-tune on UCF-101 Split 1 set.}
\vspace{-1.0cm}
\label{tab:tab9}
\end{table}

For the UCF-101 dataset, we observe that under a condition of similar spatial-temporal resolution ($T\times HW$=$8\times 224^2$), our LAPS achieves competitive accuracy (95.43\%) among 3D-CNN and Transformer pilots. With more frames and larger spatial resolutions (i.e., $32\times 320^2$, LAPS (H) further boosts accuracy to (96.85\%), reaching SOTA performance.

We test the optimal settings for the EGTEA Gaze++ dataset, where spatial-temporal resolution is $32\times 320^2$. We observe the LAPS (H) could achieve SOTA performance (66.07\%) among transformers and 3D-CNNs, verifying the robustness and generalization capacity of the LAPS model.

\begin{table}
\begin{center}
	\small
\begin{tabular}{l|lccc}
\toprule
Model &Pretrain& Res & \# Frames & EGAZ+\\
 & & ($H\times W$) &$T$  & Acc1 (\%)\\

\midrule
TSM \cite{lin2019tsm}&K-400 & $224\times224$ &8 &63.45\\
SAP \cite{wang2020symbiotic} &K-400 & $256\times256$ &64 &64.10\\
ViT (Video) \cite{dosovitskiy:image-2020}&IN-22K& $224\times224$ & 8 &62.59\\
\midrule
TokShift \cite{zhang:token-2021}&K-400& $224\times224$ & 8&{64.82}\\
TokShift-\textit{En}* \cite{zhang:token-2021}&K-400& $224\times224$ & 8&65.08\\
TokShift (HR) \cite{zhang:token-2021}& K-400& $384\times384$ &8&65.77\\
LAPS (H)&K-400& $320\times320$ & 32&\textbf{66.07}\\

\bottomrule
\end{tabular}
\end{center}
\caption{Fine-tune on EGTEA-GAZE++ Split-1 dataset. (* indicates using RGB+Optical Flow modalities)}
\vspace{-1.0cm}
\label{tab:tab8}
\end{table}
\subsection{Visualization of Exemplars with Significant Improvement}
To verify the impact of the LAPS temporal process, we compare the Base2D and LAPS model predictions on the validation set of the Kinetics-400 dataset. As video prediction is averaged over frame-level predictions (e.g, Eq. \ref{eq:vid_pred}), we present per-frame prediction of corresponding category to highlight the impact of the temporal process. We pick and compare video clips that are poorly predicted when frames are individually predicted (Base2D), but are corrected when connecting temporal relations (LAPS). Clip exemplars are shown in Figure \ref{fig:vis_all}.

We observe that: (1) By connecting long, short-term temporal relations, frame-level predictions are smoother than treating each frame individually (e.g., clips with ``\textit{Filling Eyebrows}'' and ``\textit{Skipping Rope}''), which is consistent with what we envisioned; (2) The LAPS module is more robust to failure on particular frames (e.g, 1st/5th frame of clip ``\textit{Slackling}''), and yield correct video-level prediction.

\section{Conclusion}
  We propose a zero-parameter LAPS module for video transformers. Specifically, the LAPS module is constructed from the Leap Attention and Periodic Shift sub-modules. It can flexibly replace an image transformer's generic 2D attention part with neglectable computational overhead and convert it into a video transformer. Like existing video transformers, our LAPS could not intake a very long video sequence (> 200 frames) in an end-to-end training manner, even though it is pretty efficient. But the dilated manner in temporal attention could be extended to the spatial domain to compress computations further, which will step toward long-video processing in the future.
\section*{Acknowledgments}
The work present in this paper is partially supported by the National Natural Science Foundation of China (Grant No. 62101524), by the Singapore Ministry of Education (MOE) Academic Research Fund (AcRF) Tier-1 grant, by the Zhejiang Provincial Natural Science Foundation of China (LQ21F020003), by the Exploratory Research Project (2022PG0AN01).

\vfill\eject
\bibliographystyle{ACM-Reference-Format}
\bibliography{sample-base}


\end{document}